\documentclass[sn-mathphys]{sn-jnl}
\usepackage{graphicx}
\usepackage{float}
\usepackage[section]{placeins}
\usepackage{cuted}
\usepackage{capt-of}
\usepackage{caption}
\usepackage{graphics}
\usepackage{subfigure}
\usepackage[all]{xy}
\begin{document}

\title[Toward Quantum Machine Translation]{Toward Quantum Machine Translation of Syntactically Distinct Languages}


\author[1]{\fnm{Mina} \sur{Abbaszade}}
\equalcont{These authors contributed equally to this work.}

\author*[2]{\fnm{Mariam} \sur{Zomorodi}}\email{zomorodi@pk.edu.pl}
\equalcont{These authors contributed equally to this work.}


\author*[3]{\fnm{Vahid} \sur{Salari}}\email{vahid.salari1@ucalgary.ca}

\author*[1]{\fnm{Philip} \sur{Kurian}}\email{pkurian@howard.edu}

\affil[1]{Quantum Biology Laboratory, Howard University, Washington DC, USA}

\affil[2]{Department of Computer Science, Cracow University of
Technology, Krak\'ow, Poland}

\affil[3]{Institute for Quantum Science and Technology, Department of Physics and Astronomy, University of Calgary, Alberta, Canada}


\abstract{The present study aims to explore the feasibility of language translation using quantum natural language processing algorithms on noisy intermediate-scale quantum (NISQ) devices. 
Classical methods in natural language processing (NLP) struggle with handling large-scale computations required for complex language tasks, but quantum NLP on NISQ devices holds promise in harnessing quantum parallelism and entanglement to efficiently process and analyze vast amounts of linguistic data, potentially revolutionizing NLP applications. Our research endeavors to pave the way for quantum neural machine translation, which could potentially offer advantages over classical methods in the future.
We employ Shannon entropy to demonstrate the significant role of some appropriate angles of rotation gates in the performance of parametrized quantum circuits. In particular, we utilize these angles (parameters) as a means of communication between quantum circuits of different languages.

To achieve our objective, we adopt the encoder-decoder model of classical neural networks and implement the translation task using long short-term memory (LSTM). Our experiments involved 160 samples comprising English sentences and their Persian translations. We trained the models with different optimisers implementing stochastic gradient descent (SGD) as primary and subsequently incorporating two additional optimizers in conjunction with SGD. Notably, we achieved optimal results---with mean absolute error of 0.03, mean squared error of 0.002, and 0.016 loss---by training the best model, consisting of two LSTM layers and using the Adam optimiser. 
Our small dataset, though consisting of simple synonymous sentences with word-to-word mappings, points to the utility of Shannon entropy as a figure of merit in more complex machine translation models for intricate sentence structures.
}




\maketitle

\section{Introduction}\label{sec1}
Natural language processing (NLP) refers to the use of computers for processing spoken and written language, including analyzing and understanding speech or writing produced in the format and structure of natural language, as well as producing it. Alan Turing proposed NLP in 1950 as a criterion for automated interpretation and generation of natural language, known today as the Turing test \cite{Turing}. Notably, in the context of the Turing test for CUBBITT\footnote{Charles University Block-Backtranslation-Improved Transformer Translation}, most participants struggle to distinguish CUBBITT translations from human translations \cite{Popel, Anschuetz}. In recent years, significant progress has been made in NLP with the development of advanced language models such as the Generative Pre-trained Transformer 3 (GPT-3) and its massive 175 billion parameters  \cite{GPT3} and ChatGPT with many more parameters \cite{chatGPT}. This represents a major milestone in the field of NLP, enabling the performance of tasks like machine translation, sentiment analysis, and text summarization with a level of fluency that was once thought impossible---paving the way for the development of sophisticated language models like ChatGPT that have revolutionized 
artificial intelligence (AI). As NLP continues to evolve, 
we will likely see even more impressive language models emerge in the coming years.

Moreover, the recent progress in quantum computation and information has opened up new opportunities for NLP and other fields with broad applications \cite{Lorenz, Shervin, Yulin, NRP2022, Supremacy, QML, Schuld, QML2, QML3, QML4, QML5, Havlicek, QML7, QML8, QML9, QML10, QML11}. Quantum natural language processing (QNLP) aims to define the necessary processes and mappings for NLP tasks on quantum computing devices. Specifically, new quantum approaches for NLP have been developed with the potential to outperform classical counterparts in the future \cite{f, MLST}. One potential advantage of QNLP is its ability to process large amounts of data simultaneously, which could be useful in training language models. Quantum computers can also perform certain operations faster than classical computers, which could potentially speed up the training process and lead to more accurate models. Another area where QNLP could be useful is in the analysis of context and meaning in language. Quantum systems are inherently probabilistic, which could allow for a more nuanced understanding of language and a better ability to capture the subtle nuances of meaning and context.

\begin{figure}[H]
    \centering
    \includegraphics[scale=0.4]{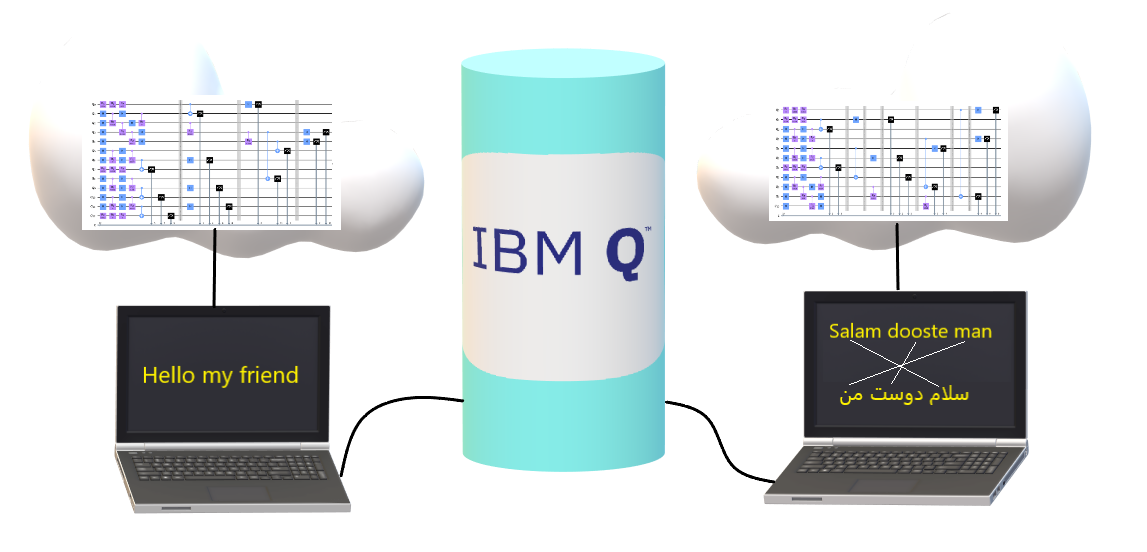}
    \caption{Quantum machine translation on noisy intermediate-scale quantum (NISQ) devices. Any syntactically distinct languages can be considered in this approach. Here is an example of a symbolic representation of translation between English and Persian on an IBM quantum computer.
    }
    \label{fig:my_label}
\end{figure}

In the QNLP model, language is interpreted as quantum processes through the use of a diagrammatic formalism in categorical quantum mechanics \cite{picturing}. QNLP protocols have two aspects, namely syntax and semantics, which are performed by mathematical operations. Semantics for quantum protocols are provided by compact closed categories \cite{protocol}. Semantics in QNLP are shown by encoding the information flow between words of the sentence. The compact closed category is used to encode this information flow. As displayed in Figure \ref{DisCoCat diagram}, cups and lines present the information flow between words of sentences.

The Distributional Compositional Categorical (DisCoCat) model of QNLP, introduced by Coecke et al. \cite{math}, utilizes a compositional grammar model, specifically the pregroup grammar, to introduce the structure of sentences. The information flow and encoded correlations between all words together in a sentence is embedded into a vector space, where vector geometry captures the correlations between words according to a given sentence. The methods for parsing sentences via pregroup grammar and drawing DisCoCat diagrams are described in various articles \cite{math, frob}.

In Ref.~\cite{inter}, the ZX-calculus is used as an easy translation between DisCoCat diagrams and quantum circuits. The ZX-calculus is a graphical language that goes beyond circuit diagrams, allowing the generalization of Pauli-Z and Pauli-X operations to show properties of circuits, entanglement states, and protocols in a visually succinct yet logically complete manner.
The ZX-calculus is a powerful tool that is shaping the next generation of quantum software. By utilizing the calculus, optimisation strategies can be employed that perform state-of-the-art T-count reduction, a key metric for fault-tolerant computing, and gate compilation. The generators of the calculus correspond closely to the basic operations of lattice surgery in the surface code, providing a visual design and verification language for these codes. Furthermore, the ZX-calculus has been used to discover novel error correction procedures, and its scalable notation allows for the representation of repeated structures at arbitrary qubit scales. 

Quantum circuit optimisation techniques based on the rules of the ZX-calculus have shown promise in simplifying ZX-diagrams \cite{Mariam}. However, this approach works best when there are few non-Clifford gates in the original circuit. By utilizing different quantum circuit optimisation techniques, the performance of quantum machine learning (QML) circuits can be improved and their cost can be reduced.

In Ref.~\cite{inter}, the use of the ZX-calculus for converting DisCoCat diagrams of sentences into quantum circuits is discussed in detail. For more information on this topic, please refer to Refs.~\cite{f}-\cite{circuit}. Additionally, Dimitri Kartsaklis et al. have developed a high-level Python library for QNLP called Lambeq \cite{lambeq}, which implements all stages of a pipeline for converting sentences to string diagrams, tensor networks, and quantum circuits ready to be used on a quantum computer.

In Ref.~\cite{Mina}, an algorithm was presented for translating QNLP tasks using long short-term memory (LSTM). Recurrent neural networks (RNNs) are commonly used in machine learning for sequence-to-sequence tasks such as machine translation \cite{thesis} and speech synthesis. RNNs can be considered as multiple copies of the same network, where each network passes a message to its successor. LSTM is a special type of RNN that can learn long-term dependencies.

In Ref.~\cite{Towards Machine translation} the systems are trained using similarity scores in basic NLP rather than deep learning models. 
The main difference between deep learning and other machine learning approaches is the way in which they learn and represent data. In traditional machine learning, such as logistic regression or decision trees, the algorithm is provided with a set of features that are engineered by humans, and the algorithm then learns to associate those features with the target output. Deep learning, on the other hand, uses a neural network architecture that is capable of automatically learning hierarchical representations of data. These neural networks are composed of many layers of interconnected nodes, and each layer learns to represent a more complex feature of the data. The deeper layers of the network learn to represent more abstract and high-level features, while the earlier layers learn to represent simpler features. This ability to learn hierarchical representations of data has led to deep learning models achieving state-of-the-art performance in many machine learning tasks, including NLP, computer vision, and speech recognition. In summary, the key difference between deep learning and other machine learning approaches is the way in which they learn and represent data. Deep learning models use a hierarchical neural network architecture to learn complex representations of data, which can lead to superior performance in many machine learning tasks.

There are several approaches to the concept of quantum machine translation in the academic literature. For instance, one study posits the implementation of a quantum neural network-based machine translator that is specifically designed for English language translations \cite{R}. Another scholarly article serves as an introductory guide to the field of quantum machine learning, which has significant implications for applications such as image and speech recognition or strategic optimisation \cite{M.Schuld, Schuld}. Another approach outlines the development of a deep learning system that is optimised for machine translation \cite{Popel}. Additionally, a systematic review of existing literature pertaining to quantum machine learning is available \cite{Peral, Caro}. Lastly, QNLP has been developed to represent the meaning of phrases as vectors encoded in quantum computers.

In our work, an LSTM model that has shown good performance on non-trivial tasks such as sequence learning was constructed. To feed the model with inputs, the quantum circuit was encoded into a multidimensional space of vectors that corresponded to the embedding of the initial words. This encoding was used to train an LSTM model, which was later decoded to the corresponding quantum circuits.

The paper is structured as follows: 
Section \ref{methods} explains a summary of all the methods used in the paper.
Section \ref{quantum circuit} describes the DisCoCat diagram and how it is converted into a quantum circuit. In this section, a dataset of sentences from two different languages is provided, and the Shannon entropies of their quantum circuits are evaluated. Section \ref{implementation} provides an implementation for the translation task of QNLP via a classical LSTM model. In this section, we describe how a dataset of 160 sentences in English and Persian was prepared, and how the parameterized quantum circuit of similar sentences in English and Persian was designed.

\section{Methods}\label{methods}

We utilized DisCoCat diagrams to represent sentences from two syntactically distinct languages (English and Persian). Through the application of the specified operations, these diagrams were transformed into quantum circuits. A dataset of quantum circuits was then compiled, containing both Persian and English sentence circuits. The English sentence circuits were separated into two groups, with one group having identical angles of rotation gates for synonymous words in English and Persian sentences. The angles of rotation gates were then altered to create another group of English quantum circuits, where synonymous words between English and Persian sentences did not share the same angles for their rotation gates. Subsequently, the quantum circuits were simulated for 20,000 shots on the IBMQ-qasm-simulator, and the resulting frequency distributions were used to generate probability distributions; i.e., after running a quantum circuit, the occurrence frequency of each state is divided by the sum of all occurrence frequencies to generate the probability of the state.

We propose a novel method for the purpose of aligning
source and target sentences. Our method involves running quantum circuits on both the source and target sentences, and subsequently evaluating the Shannon entropy from the output probability distributions of the resulting circuits. We then measure the difference between the Shannon entropy values of each pair of source and target sentences. Our method has yielded an almost regular pattern of differences, which is expected to guide quantum neural machine translation in avoiding the training of obviously incorrect sentences in explicit direct mapping models. Moving forward, we intend to add a new layer to recurrent quantum neural networks using this approach, which can be applied to various datasets to ensure their synchronization. This approach has also been implemented in classical neural networks \cite{match}.
Moreover, we created a dataset consisting of 160 quantum circuits for English and Persian sentences, with synonymous words from both languages having identical parameters. By utilizing classical LSTM, we were able to perform the translation task via the encoding-decoding of quantum circuits of sentences.

\section{Quantum circuits of sentences in two different languages and evaluating Shannon entropy}\label{quantum circuit}

The aim of QNLP is to create a framework that captures both semantic and syntactic aspects of sentences using a mathematical form known as the compact closed category. Quantum maps are introduced to represent meaning in QNLP and developed using a diagrammatic language for representing processes and their compositions \cite{p,Grefenstette}. This non-commutative categorical quantum logic language allows for the reduction of diagrams of sentences, and facilitates comparison of grammatical structures in different languages. Pregroups are utilized to encode the grammar of languages, providing an algebraic analysis of sentence parsing \cite{m}. Models for the semantics of positive and negative transitive sentences and relative pronouns are presented in Refs.~\cite{math, frob}.

\subsection{Conversion of sentences to quantum circuits}
In this section, we will summarize converting a sentence into a quantum circuit by a mathematical framework, with complete explanations in Ref. \cite{Mina}. Vector spaces and pregroups are used to assign meanings to words and grammatical structures to sentences in a language, which can then be converted into quantum circuits.
A compact closed category is a monoidal category where for each object $A$ there are objects $A^r$ and $A^l$, and morphisms
	\begin{align*}
	& \eta ^l : I \rightarrow A\otimes A^l,~~~~ 
	\eta ^r : I \rightarrow A^r \otimes A \\
	&\epsilon^l : A^l \otimes A \rightarrow I,~~~~ 
	\epsilon^r : A \otimes A^r \rightarrow I.
	\end{align*}
 
 In the graphical language the $\eta$ maps are depicted by caps, and $\epsilon $ maps are depicted by cups \cite{math}. 
 A pregroup is a partially ordered monoid 
 $(P, \leqslant, \cdot, 1)$
 with multiplication monoid ‘$\cdot$’ and unit ‘$1$’.
 For each  $p\in P$ there are elements $p^l, p^r \in P $ such that
 $$ p ^l \cdot p \leqslant 1 \leqslant p \cdot p^l, ~~~
  p \cdot p^r \leqslant 1 \leqslant p^r \cdot p$$
  
Pregroup is a compact closed category. Morphisms are reductions and the operation ‘$\cdot$’ is the monoidal tensor of the monoidal category. 
In order to transition from classical to quantum computing in QNLP, each sentence is encoded as a quantum circuit to be executed on a quantum computer. 
In this process, we follow the steps outlined below:

Initially, we select synonymous sentences in English and Persian. As an example, we consider the English sentence ‘‘Sara buys the book from the bookshop" and its Persian counterpart ‘‘Sara ketab ra az ketabforoushi mikharad''. Next, we utilize pregroup grammar to analyze and parse the sentences \cite{math}. 
Let the basic types be:\\
\\
$n$ : noun,\\
$s$ : declarative statement\\ 
\\
and generate the free pregroup of this type \cite{Mina}. If the types of words in a sentence are reduced to the basic type ``s'' through juxtaposition, the sentence is considered grammatical.

It is important to note that the relationship $n^l . n \leqslant 1$ holds, which implies the existence of a morphism $\epsilon^l : n^l \otimes n \rightarrow I$ within a compact closed category, with symbols representing the various components. 

\begin{figure}[H]
\centering
\includegraphics[scale=0.65]{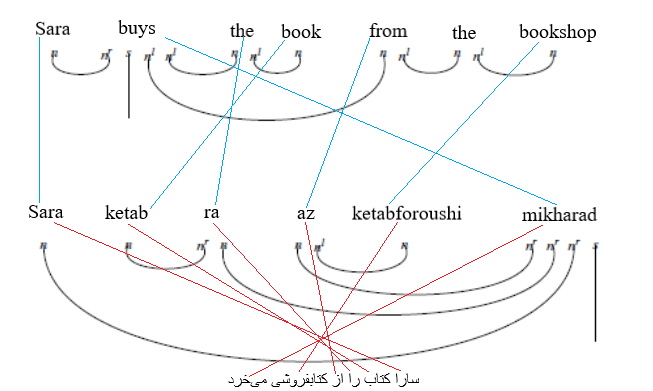}
\caption{Top and bottom diagrams represent DisCoCat diagrams of English and Persian, ``Sara buys the book from the bookshop'' and ``Sara ketab ra az ketabforoushi mikharad,'' respectively.}
\label{DisCoCat diagram}
\end{figure}

 The assignment of types to words within a sentence is determined by their grammatical position. In the example ``Sara buys the book from the bookshop'', we connect all words of the sentence to the verb ``buys''. Let's consider the words ``the'' and ``bookshop'' as an example. In this case, we need to establish a connection between the word ``bookshop'' and ``the''. To represent the noun ``bookshop'' we assign the type ``n'', and to connect it to the word ``the'', we use $n^l$. Both ``the'' and ``bookshop'' should also be connected to the word ``from''. Consequently, we assign ``n'' to the word ``the'' and adjoin it to ``from'' using $n^l$. To connect ``Sara'' to the verb, we put the right-adjoint $n^r$ to ``buys''. Similarly, by assigning ``n'' to the word ``from'', we adjoin it to the verb ``buys'' using $n^l$. This process continues for all words in the sentence. These assignments are not unique for words, and they change according to the structure of the sentences. For more comprehensive information, please refer to \cite{algebraicapproach}. 
 According to \cite{algebraicapproach}, the example sentence ``Sara buys the book from the bookshop" has the following type of assignment: 

\vspace{.5cm}
\begin{center}
	Sara \ \ \ buys \ \ \ the \ book \ from  \ \ \ the \ \ \ bookshop.
	\\
	$ \hspace{-.7cm} n \hspace{.4cm} (n^r s n^l n^l) \hspace{.1cm} (n n^l) \ \ \  n \hspace{.4cm} (n n^l) \ \ \ (n n^l) \ \ \ \ \ \  n $
\end{center}
\vspace{.5cm}
This is grammatical because of the following reduction: 
\vspace{.5cm}
\begin{center}
	$ (n n^r) s n^l (n^l n) (n^l n) n (n^l n)(n^l n) \rightarrow s (n^l n)  \rightarrow s $.
\end{center}
\vspace{.5cm}
According to \cite{m}, the example sentence ``Sara ketab ra az ketabforoushi mikharad" has the following type of assignment: 
\begin{center}
	Sara \ \ ketab \ ra \ \ \ \ az \ ketabforoushi mikharad.
	\\
	$ n \hspace{.9cm} n \hspace{.5cm} (n^r n)\hspace{.2cm} ( n n^l) \ \ \ \ \ n \hspace{1cm} (n^r n^r n^r s)$
\end{center}
\vspace{.5cm}
This is grammatical because of the following reduction: 
\vspace{.5cm}
\begin{center}
	$n (n n^r) n n (n^l n)  n^r n^r n^r s\rightarrow n n (n  n^r) n^r n^r s \rightarrow n (n n^r) n^r s \rightarrow (n n^r) s \rightarrow s $.
\end{center}
\vspace{.5cm}
The reductions and types are interpreted as linear maps and vector spaces, obtained by a monoidal functor from pregroup to finite dimensional vector spaces. For the above sentences, the reductions are depicted diagrammatically in Figure \ref{DisCoCat diagram}, in which each cup has two wires, i.e. it has two inputs. For more details see \cite{Mina}.

Now by using the ZX-calculus we translate DisCoCat diagrams into quantum circuits. We consider one qubit for every wire of type $n$ and $s$. According to \cite{math}, wires are the input and output of morphisms.

According to the definition of a pregroup, $n^l$ and $n^r$ are of type $n$, named left and right adjoints of $n$. So the verb is a state of some qubits. Figure \ref{DisCoCat diagram} shows that the quantum circuit for the English sentence consists of 13 qubits, 1 for ``Sara'', 4 for ``buys'', 2 for ``the'', 1 for ``book'', 2 for ``from'', and 1 for ``bookshop''. The quantum circuit for the Persian sentence consists of 11 qubits, 1 for ``Sara'', 1 for ``ketab", 2 for ``ra'', 2 for ``az'', 1 for ``ketabforoushi'', and 4 for ``mikharad''. 
This comes directly from the fact we have assigned 1 qubit to noun wires and 1 qubit to sentence wires. DisCoCat diagrams of sentences are constructed based on reduction maps and the representation of $\epsilon$ maps using cups. In these diagrams, the number of connections to each word corresponds to the number of wires present.

Each noun qubit corresponds to two R$_x$ rotations and one R$_z$ rotation, which together implement an Euler decomposition to determine the qubit's state. The four verb qubits must be connected, which is achieved using instantaneous quantum polynomial (IQP) layers consisting of Hadamard gates and diagonal unitaries. In this specific circuit, a single IQP layer is used, where all qubits receive Hadamard gates and the 1st and 2nd, 2nd and 3rd, and 3rd and 4th qubits are connected by a conditional R$_z$ gate (a diagonal unitary), resulting in all four qubits being linked to represent the state of the English verb ``buys''. This method is applied similarly to Persian sentences. The theory behind IQP circuits and why they are a good choice for QML is explained in detail in the paper \cite{Havlicek}. The cups (tensor contractions in standard DisCoCat) correspond to Bell effects and require post-selection where we measure all qubits many times and keep the shots where 12 out of the 13 qubits (excluding the sentence qubit) are zero. It is possible to have more than one IQP layer.

\begin{figure}[H]
\centering
\includegraphics[width=0.9\textwidth]{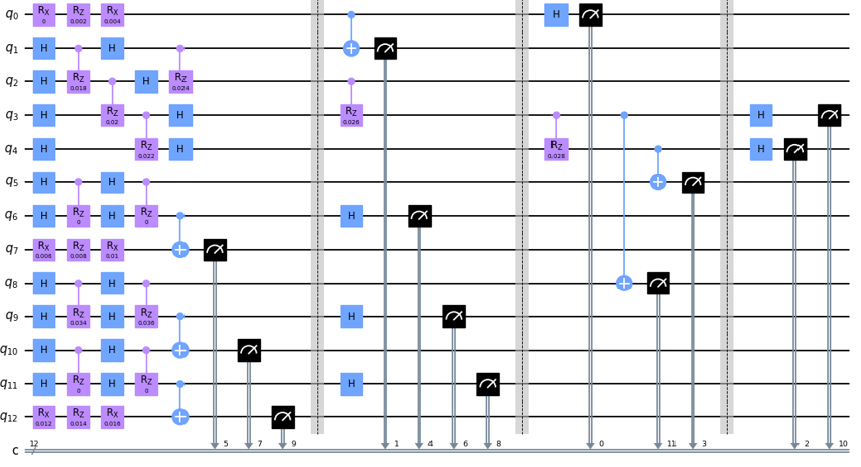}
\caption{The quantum circuit of ``Sara buys the book from the bookshop'' in English.}
\label{English}
\end{figure}

\begin{figure}[H]
\centering
\includegraphics[width=0.9\textwidth]{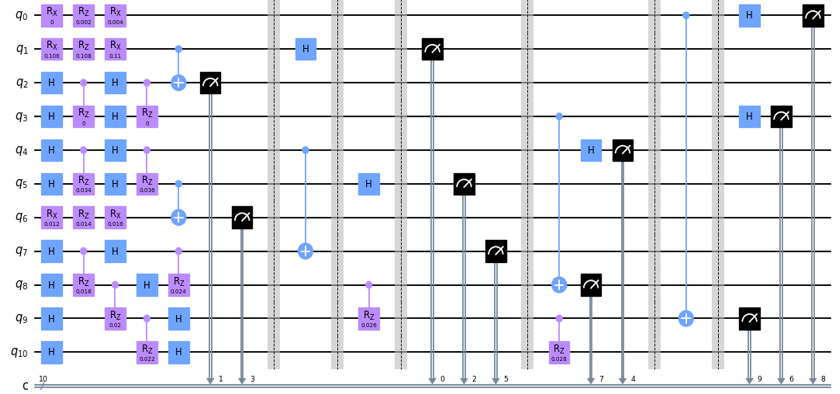}
\caption{The quantum circuit of ``Sara ketab ra az ketabforoushi mikharad" in Persian.}
\label{Persian}
\end{figure}

In Qiskit format, Fig \ref{English} and Fig \ref{Persian} represent the quantum circuit of the sentence ``Sara buys the book from the bookshop" in English and in Persian, respectively. We identify the same angles for rotation and controlled rotation gates for synonymous words in each pair of quantum circuits for English and Persian sentences. 
We select the synonymous keywords that are the same between sentences of English and Persian, and put the same angles for their rotation gates. Other words can take random angles for their rotation gates. Because two syntactically distinct languages will not have a one-to-one mapping between corresponding words,
future work will entail how multiple-word syntactic constructions in one language map to various multiple-word alternatives in another. The structure of complex phrases in English \cite{frob} and the rules of more complex parsing protocols \cite{lambeq} could be used to distinguish assignment patterns for basic types and morphisms between noun-centric (thing-oriented) and verb-centric (process-oriented) languages.

\subsection{Dataset creation for evaluating Shannon entropy}
Shannon entropy is a widely used measure that quantifies the level of statistical uncertainty or randomness in a system. Derived from the Boltzmann entropy in the microcanonical ensemble, it has proven to be particularly useful in assessing the amount of information present in a given signal or dataset. In the context of machine learning algorithms, Shannon entropy is a useful tool to measure the dissimilarity or distance between different samples or clusters of data. 
Alternatively, Shannon entropy can be used to measure distance by computing the entropy of each sample or cluster and comparing it to the entropy of the overall dataset. However, care should be taken: Consider the case where there is only one state in each distribution. The probability of that state is one, and so the Shannon entropy is zero for each distribution. However, two such states could be separated by a very large distance (e.g., orthogonal ``0'' and ``1'' signals). A much ``closer'' state to either distribution could be constructed, which has more Shannon entropy but is much closer to the signal in the one-state distribution with zero entropy. This should be addressed with care for each use case and identified as a possible source of error in analysis.

Furthermore, Shannon entropy can also be applied to feature selection and dimensionality reduction in machine learning. The entropy of each feature can be calculated and the feature with the highest information 
can be chosen as the best representation of the dataset.
It is important to recognize that while Shannon entropy is a valuable metric for measuring differences in machine learning algorithms, its effectiveness is contingent upon the unique characteristics of the dataset and the problem at hand. 
Although our study focuses on English and Persian languages, it is important to acknowledge that other languages may also be applicable. Table \ref{1} displays several example sentences from the aforementioned languages, and each sentence was run through a quantum circuit and evaluated using Shannon entropy on the IBMQ-qasm-simulator with 20,000 shots. 
After running a quantum circuit that contains $n$ qubits, we get $2^n$ different states and their frequencies. The frequency of each state represents the number of occurrences of that state after 20000 shots. By dividing the frequency of each state $x_i$ by 20000, we get the probability of that state as $p(x_i)$.
Let $X$ represent the probability distribution of running the quantum circuit for a given sentence.
The Shannon entropy $H$ is calculated as follows:
$$ H(X) = - \sum\limits_{i \in I} p(x_i) \log p(x_i)$$
For each state $ x_i \in X,$  $p(x_i)$ is the probability of $x_i$. $I$ is the index set that represents the $2^n$ elements of $X$. This means that $I$ represents the number of states observed in the circuit execution results file for a sentence.

\begin{figure}[H]
\centering
\includegraphics[width=0.9\textwidth]{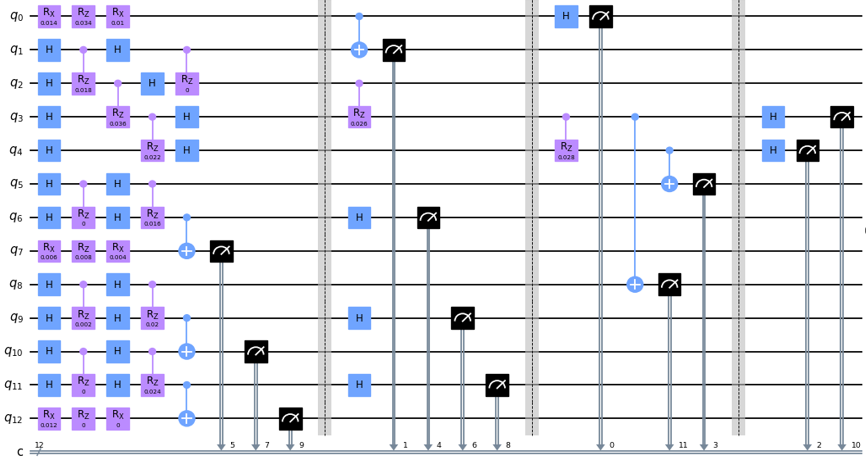}
\caption{The quantum circuit of ``Sara buys the book from the bookshop'' after swapping angles of rotation gates of Fig \ref{English}. We maintain all gates and the overall structure of the circuit.}
\label{ch English}
\end{figure}

In this research, we put the same angles for synonymous words in quantum circuits of English and Persian sentences, and evaluate the Shannon entropy of each quantum circuit. Next, we proceed to swap the angles of rotation gates of each quantum circuit for the English sentence randomly,
and repeat the experiment, ensuring that the angles corresponding to synonymous words in the English and Persian sentences are not the same. This is done in order to maintain the original angles of each circuit as much as possible. In other words, the angles of rotation gates of a quantum circuit for each English sentence are swapped and the experiment is repeated. An illustrative example of this process can be seen in Figure \ref{ch English}. The new angle of each rotation gate is selected from the original angles, randomly. For example, for the two sentences -- ‘Sara buys the book from the bookshop’ in English and  ‘Sara ketab ra az ketabforoushi mikharad’ in Persian/Fenglish\footnote{``Fenglish" is derived from translation of the words in Persian (i.e., ``Farsi''), using the English alphabet instead of the Persian alphabet.} -- the synonymous words are:\\
Sara $\equiv$ Sara\\
buys $\equiv$ mikharad\\
book $\equiv$ ketab\\
bookshop $\equiv$ ketabforoushi\\
from $\equiv$ az\\
For the above synonymous words, first, we put the same angles for their rotation gates presented in Figures \ref{English} and \ref{Persian}. Then we swapped angles in each English circuit and repeated the experiment.

\begin{table}[htbp]
\caption{The table below presents synonymous sentences in two languages, English and Persian. The last column shows the Persian translation of the sentences in Latin letters, referred to as ``Fenglish''.}\label{1}
\includegraphics[width=1.0\textwidth]{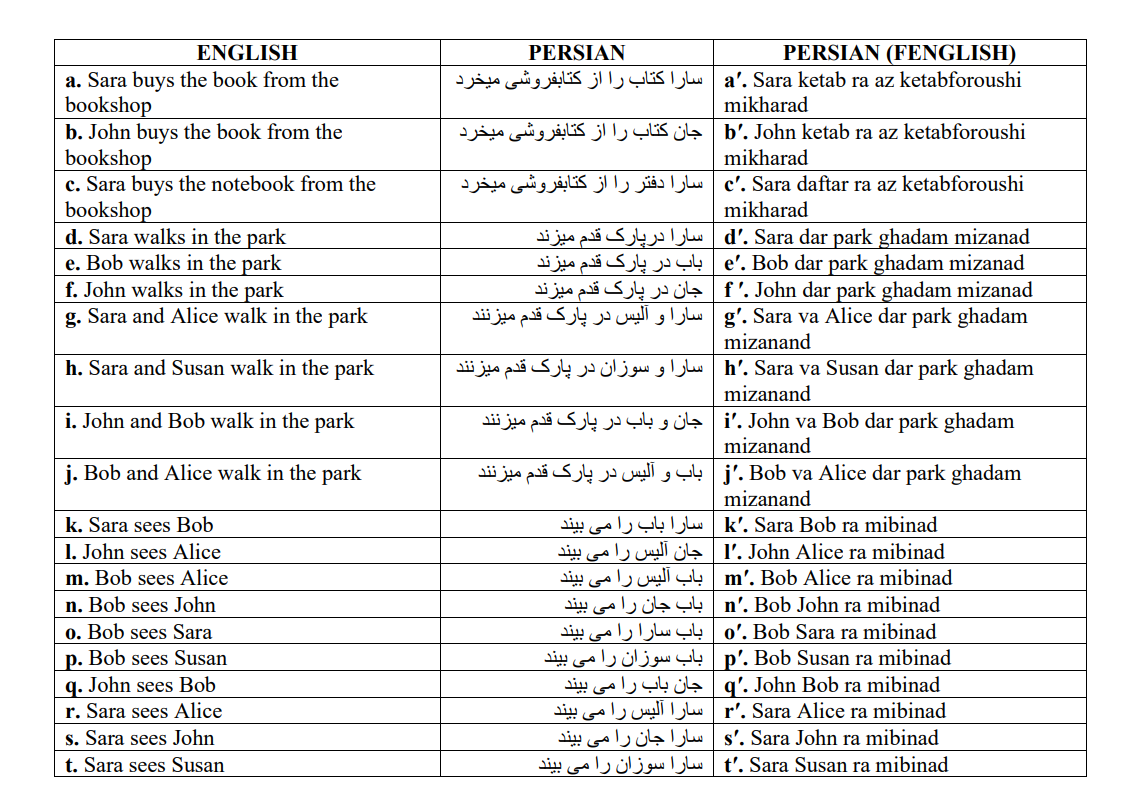}
\end{table}

\subsection{The results for evaluating Shannon entropy of quantum circuits of sentences}
\begin{figure}[H]
\centering
\includegraphics[width=0.9\textwidth]{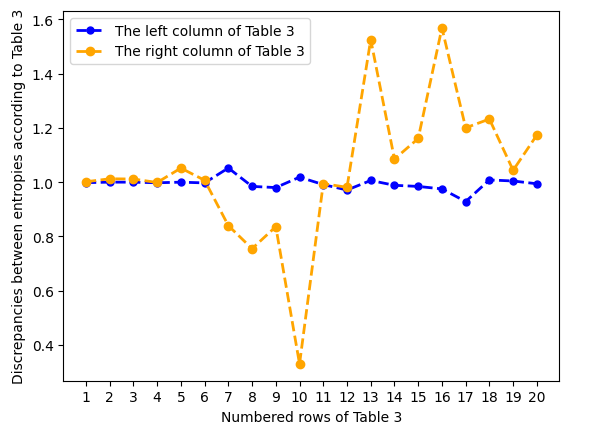}
\caption{The left and right columns of Table \ref{difference} represent the discrepancies of Shannon entropy of quantum circuits of English sentence and its translation in Persian that shared the identical and not-identical angles in quantum circuits.}
\label{diagram shanon entropy}
\end{figure}

\begin{figure}[htp]
\centering
\includegraphics[width=0.9\textwidth]{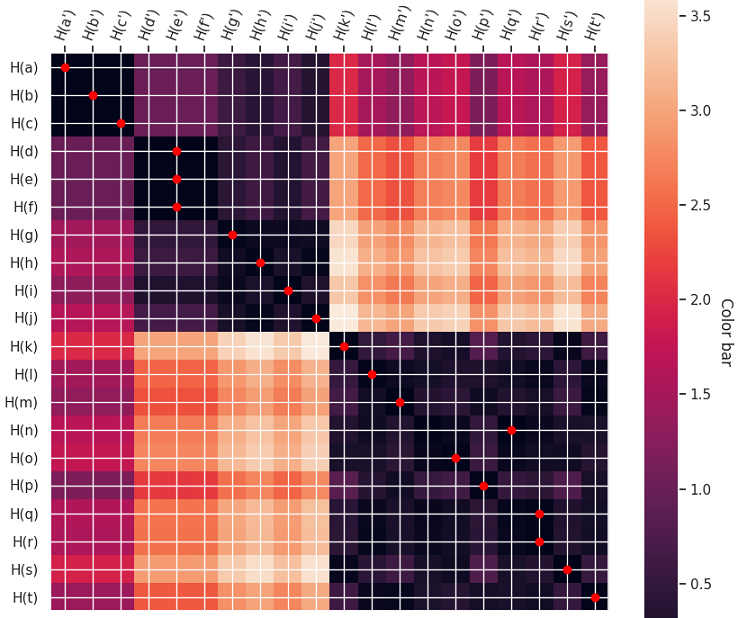}
\captionof{figure}{The heat map representation of Shannon entropy differences between syntactically distinct languages (English and Persian) for finding the best translation for each sentence in a database. The heat map displays the absolute value of
the pairwise entropy differences (from the left and middle columns of Table 2) minus unity. Based on this representation, the entropy differences closest to unity are in the dark regions mainly along the diagonal elements, shown by red dots on the heat map as the best match for translation. See main text for further details.}
\label{heat map}
\end{figure}

Shannon entropy metric may prove to be a valuable tool in analyzing QNLP data.
In this section, we have conducted a thorough analysis of the quantum circuits of sentences, which involved two primary steps. We conducted experiments using quantum circuits to compare synonymous sentences in English and in Persian. 
First, to ensure consistency, in each quantum circuit we used similar angles of rotation gates for words with similar meanings in each sentence.
Second, we adjusted the angles of rotation gates in quantum circuits of English to ensure that synonymous words in both languages did not share the same angles. Thus, we sought for a way to match each pair of English and Persian sentences and add a matching layer in the recurrent neural network.
The resulting circuits were executed using the IBMQ-qasm-simulator with 20000 shots, and the corresponding Shannon entropy values for each circuit are presented in Table \ref{Shannon entropy}. The third column of Table \ref{Shannon entropy} depicts the Shannon entropy values for English sentences with angles exchanged, while Table \ref{difference} shows the discrepancies between the first and second, as well as the second and third columns of Table \ref{Shannon entropy}. 
The results of this analysis are presented in Figure \ref{diagram shanon entropy}. 
Second, we utilized the left and middle columns of Table \ref{Shannon entropy}, which assigned similar angles of rotation gates to the synonymous words in quantum circuits of English and Persian sentences, and calculated the Shannon entropy of each quantum circuit of all English and Persian sentences listed in Table \ref{1}. The results of this analysis are presented in Figure \ref{heat map}. 

\begin{sidewaystable}[htbp]
\sidewaystablefn%
\begin{center}
\caption{The results of evaluating the Shannon entropy of quantum circuits for English and Persian sentences are shown in the left and middle columns, respectively. In both cases, synonymous words in the sentences were assigned the same angles in quantum circuits. The right column displays the results of evaluating the Shannon entropy of quantum circuits for English sentences after angle swapping.}\label{Shannon entropy} 
\begin{tabular}{|l|l|l|}
\hline
\bfseries{Shannon entropy of quantum} &  \bfseries{Shannon entropy of quantum} &
\bfseries{Shannon entropy of quantum}\\
\bfseries{circuits of English sentences} &  \bfseries{circuits of Persian sentences} & \bfseries{circuits of English sentences} \\
\hline
$H(a)= 6.001315160873559 $ & $H(a')= 5.0031967710951015 $ & $H(a^{\prime \prime})= 6.004736428636913$\\
\hline
$H(b) = 6.006578931000881 $ & $H(b') = 5.006429642328229 $ & $H(b^{\prime \prime})= 6.018957659399867$\\
\hline
$H(c) = 6.005966787750403 $ & $H(c') = 5.005974227809354 $ & $H(c^{\prime \prime})= 6.017595798609497 $\\
\hline
$H(d) = 6.995594756756778 $ & $H(d')= 5.997804218054402 $ & $H(d^{\prime \prime})= 6.996647874624478 $\\ 
\hline
$H(e) = 6.995539523537989 $ & $H(e')= 5.99550229119629  $ & $H(e^{\prime \prime})= 7.04742881564586 $\\
\hline
$H(f) = 6.995713266260664 $ & $H(f')= 5.998383911173536  $ & $H(f^{\prime \prime})= 7.005377911621739 $\\
\hline
$H(g) = 7.472076459921503 $ & $H(g')= 6.419834213204461 $ & $H(g^{\prime \prime})= 7.259668533783197 $\\
\hline
$H(h) = 7.58181670756182 $ & $H(h')= 6.59739502901938 $ & $H(h^{\prime \prime})= 7.352212286404398 $\\
\hline
$H(i) = 7.315157675950335  $ & $H(i')= 6.335068026209578 $ & $H(i^{\prime \prime})= 7.171110796325161 $\\
\hline
$H(j) = 7.6592276186465345  $ & $H(j')= 6.640685720752816 $ & $H(j^{\prime \prime})= 7.3294011238656 $\\
\hline
$H(k) = 2.0006784957219317 $ & $H(k')= 2.999789504246745 $ & $H(k^{\prime \prime})= 2.0049399527709584$\\ 
\hline
$H(l) = 2.5203329918706316 $ & $H(l')= 3.4914554204229953 $ & $H(l^{\prime \prime})= 2.510412409090681$\\ 
\hline
$H(m) = 2.6645482358797676 $ & $H(m')= 3.6707868709904896  $ & $H(m^{\prime \prime})= 2.1451431789662303 $\\
\hline
$H(n) = 2.330285405108207 $ & $H(n')= 3.3187826378940786  $ & $H(n^{\prime \prime})= 2.2337291491135707 $\\
\hline
$H(o) = 2.2449753288648266 $ & $H(o')= 3.2293084705194266  $ & $H(o^{\prime \prime})= 2.067748117020039 $\\
\hline
$H(p) = 2.8533918863400323 $ & $H(p')= 3.828552220833362  $ & $H(p^{\prime \prime})= 2.259740814655625 $\\
\hline
$H(q) = 2.4035414027599233 $ & $H(q')= 3.332105656985817  $ & $H(q^{\prime \prime})= 2.131743480610954 $\\
\hline
$H(r) = 2.4224007881842406 $ & $H(r')= 3.430680111906539 $ & $H(r^{\prime \prime})= 2.19789321159273 $\\
\hline
$H(s) = 2.0807634799644323 $ & $H(s')= 3.08513013099599 $ & $H(s^{\prime \prime})= 2.0414209912678674 $\\
\hline
$H(t) = 2.602246530030625 $ & $H(t')= 3.611661842800734 $ & $H(t^{\prime \prime})= 2.437846467686803 $\\
\hline
\end{tabular}
\end{center}
\end{sidewaystable}

\begin{table}[htbp]
\begin{center}
\caption{The left column shows discrepancies between the left and the middle columns of Table \ref{Shannon entropy}. The right column shows discrepancies between the middle and the right columns of Table \ref{Shannon entropy}.}\label{difference}
\begin{tabular}{|l|l|}
\hline
\bfseries{discrepancies between entropies} &  \bfseries{discrepancies between entropies} \\
\hline
$ \vert H(a)-H(a') \vert = 0.998118389778458 $  &  $ \vert H(a^{\prime \prime})-H(a')\vert= 1.001539657541812 $ \\
\hline
$ \vert H(b)-H(b')\vert = 1.000149288672652 $ & $ \vert H(b^{\prime \prime})-H(b')\vert = 1.012528017071638 $\\
\hline
$ \vert H(c)-H(c')\vert = 0.999992559941049 $ & $ \vert H(c^{\prime \prime})-H(c')\vert = 1.011621570800143 $\\
\hline
$ \vert H(d)-H(d')\vert = 0.997790538702376 $ & $ \vert H(d^{\prime \prime})-H(d')\vert= 0.998843656570076 $\\ 
\hline
$ \vert H(e)-H(e')\vert = 1.00003723234167 $ & $ \vert H(e^{\prime \prime})-H(e')\vert = 1.05192652444957 $\\
\hline
$ \vert H(f)-H(f')\vert = 0.997329355087128 $ & $ \vert H(f^{\prime \prime})-H(f')\vert = 1.00699400044820 $\\
\hline
$ \vert H(g)-H(g')\vert = 1.052242246717042 $ & $ \vert H(g^{\prime \prime})-H(g')\vert= 0.839834320578736 $\\
\hline
$ \vert H(h)-H(h')\vert = 0.98442167854244 $ & $ \vert H(h^{\prime \prime})-H(h')\vert = 0.75481725738508 $ \\
\hline
$ \vert H(i)-H(i')\vert = 0.980089649740757 $ & $ \vert H(i^{\prime \prime})-H(i')\vert = 0.836042770115583 $\\
\hline
$ \vert H(j)-H(j')\vert = 1.018541897893724 $ & $ \vert H(j^{\prime \prime})-H(j')\vert = 0.329826494780934 $\\
\hline
$ \vert H(k)-H(k')\vert = 0.9911100852481 $ & $ \vert H(k^{\prime \prime})-H(k')\vert = 0.994849551475787 $\\
\hline
$ \vert H(l)-H(l')\vert = 0.971122428552364 $ & $ \vert H(l^{\prime \prime})-H(l')\vert = 0.98104311332314 $\\
\hline
$ \vert H(m)-H(m')\vert = 1.006238635110722 $ & $ \vert H(m^{\prime \prime})-H(m')\vert = 1.52564369202425 $\\
\hline
$ \vert H(n)-H(n')\vert = 0.988497232785871 $ & $ \vert H(n^{\prime \prime})-H(n')\vert = 1.085053488780508 $\\
\hline
$ \vert H(o)-H(o')\vert = 0.9843331416546 $ & $ \vert H(o^{\prime \prime})-H(o')\vert = 1.1611560353499396 $\\
\hline
$ \vert H(p)-H(p')\vert = 0.97516033449333 $ & $ \vert H(p^{\prime \prime})-H(p')\vert = 1.568811406177742 $\\
\hline
$ \vert H(q)-H(q')\vert = 0.928564254225894 $ & $ \vert H(q^{\prime \prime})-H(q')\vert = 1.200362176374867 $\\
\hline
$ \vert H(r)-H(r')\vert = 1.00827932372229 $ & $ \vert H(r^{\prime \prime})-H(r')\vert = 1.23278690031383 $\\
\hline
$ \vert H(s)-H(s')\vert = 1.004366651031468 $ & $ \vert H(s^{\prime \prime})-H(s')\vert = 1.043709139728033 $\\
\hline
 $\vert H(t)-H(t')\vert = 0.9941531277011$ & $\vert H(t^{\prime \prime})-H(t')\vert = 1.17381537511393$ \\
\hline
\end{tabular}
\end{center}
\end{table} 

For the results that are illustrated in Figure \ref{diagram shanon entropy}, the blue line represents the differences in Shannon entropy between English and Persian circuits with the same angles of rotation gates for synonymous words, while the orange line represents the discrepancies for circuits with different angles of rotation gates for synonymous words. The blue line exhibits a more regular pattern than the orange line. It is important to note that noise and errors in quantum computing can impact the accuracy of these results, which is particularly relevant for applications such as cryptography or chemical simulations that require high precision and reliability.
Figure \ref{heat map} represents the heat map matrix representation of the pairwise entropy differences between all quantum circuits of sentences in Table \ref{1}.\footnote{The numerical values for the $20 \times 20$ matrix elements representing the entropy differences between all quantum circuits in Table \ref{1} are made available via access link in Section \ref{Data}.}
The matrix represents differences between the first and second columns of Table \ref{diagram shanon entropy}. Entries on the main diagonal of the matrix are the values closest to unity. Figure \ref{heat map} thus represents the absolute value of the pairwise entropy differences, minus unity. We have marked the location of these numbers in red dots as the best matches. By putting the same angles for rotation and controlled rotation gates for synonymous words in English and Persian sentences, according to Figure \ref{heat map}, the distance between each English sentence and its translation in Persian is close to unity. Where the best matches are not on the diagonal, one can see by inspection that these cases arise from very similar syntactic structures between sentences (e.g., `Sara walks in the park' vs. `Bob walks in the park').

Noise can impact the outcomes obtained from quantum circuits executed on IBM quantum computers. Decoherence, the quantum state disturbance or loss of a qubit due to interaction with the environment, is one of the major sources of noise in quantum computers. This can lead to errors in the quantum state and, consequently, impact the overall computation. Another source of noise is control errors, which arise during qubit manipulation, such as during gate operations, and can also impact the quantum state and the computation results. However, since noise and errors are inherent in quantum computing, it is essential to acknowledge their limitations. Therefore, the precision in our quantum approach is affected by this restriction and will be improved by future generations of NISQ devices with significantly lower error rates. 

\section{Implementing the translation task via QNLP}\label{implementation}

Our test dataset in this section includes 160 sentences (augmented from the 20 sentences in the previous section), 80 sentences for each language in the translation task.\footnote{The access links to this dataset are available in Section \ref{Data}.} We considered the same angles of rotation gates for the synonymous words in quantum circuits of English and Persian sentences. 
 
\begin{figure}[H]
\centering
\includegraphics[width=4.9in]{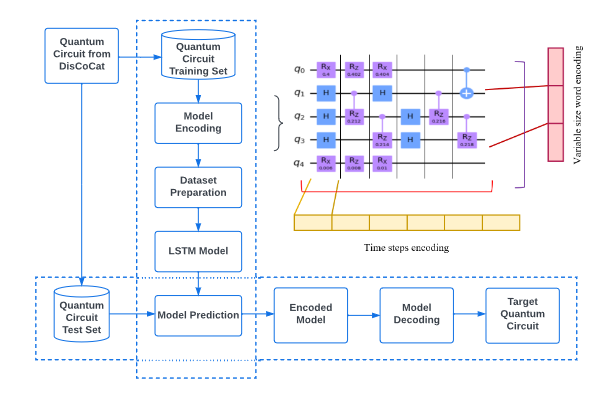}
\caption{Steps of the proposed method for the hybrid quantum-classical machine translation.}
\label{QMT}
\end{figure}

We have a classical computer interface and the system works as a quantum-classical hybrid. The input to the machine translation network comes from a decoded quantum circuit appropriate for the network. Then the result is converted back to a quantum circuit according to the encoding predicted by the network. 
Figure \ref{QMT} shows the steps of the proposed method for encoding the quantum circuit into the LSTM input shape and then converting it back into the output quantum circuit in the target language.

\subsection{Quantum circuit encoding for translation task\label{Encoder-Decoder}}
To train our LSTM model using quantum circuits, we must first encode the circuits into a format that can be understood by the model and decode them back into the appropriate quantum circuit. This encoding should follow the same principles that feed sentences in a given language. But this time, other than having the encoding of a sentence conventionally, the input is encoded from a quantum circuit. The result of encoding is represented as a nested vector of length $\mathbb{L}$ corresponding to the same number of words in the quantum circuit encoding. Our dataset $D$ consists of source and target language sentence pairs, and we encode the set of quantum circuits $QC$ from $D$. Each pair of quantum circuits ${qc}{s}, {qc}{t} \in QC$ represents mappings for sentences in different languages using DisCoCat diagrams. For each quantum circuit in $QC$, its architecture is extracted and encoded as a vector $E$. The encoding $E$ is constructed for the given DisCoCat-induced quantum circuit in a space corresponding to the quantum circuit model and is fed to the classical LSTM network.
The network reads the quantum circuit from the first time step $t_0$ to the last time step $t_{n-1}$ for every word in the sentence encoded from the quantum circuit.
Each word has different time steps, each time step composed of a series of quantum gates, and each quantum gate has its own encoding. Therefore, we build a nested structure to prepare the encoding.
Therefore the encoding of each word $w_l$ is constructed as a list of quantum gates in different time steps and is represented by $E_{w_{l}}(t_{w_l}^{i}(g_k))$. 

\vspace{.5cm}
\begin{center}
 $ E_{w_l} = [t_{w_l}^0, t_{w_l}^1, \cdots, t_{w_l}^{n-1}]$
\end{center}
\vspace{.5cm}
Here, $E_{w_l}$ is a vector of time steps for word $w_l$ in the quantum circuit. This vector has the same length for all words. Each element of this vector is represented as:
\vspace{.5cm}
\begin{center}
  $ t_{w_l}^i = [[{g_1}^i], [{g_2}^i], \cdots, [{g_k}^i]] $
\end{center}
\vspace{.5cm}
$t_{w_l}^i$ shows gates inside word $w_l$ for time step $i$. The left-hand side of the equation above is the symbol to represent the $i$th time step of the $l$th word in the sentence, and the right-hand side shows the vector of this word, which consists of $k$ gates.

We represent each element in $t_{w_l}^i$ by a vector as above for $ 1 \leq j \leq k $, where $k$ is the number of gates in time step $i$. $k$ is not constant for each word, so we used padding to make them of equal size.
$[{g_j}^i]$ is the vector representation for the gate $g_j$ in time step $i$ of word $w_l$. We used index $j$ to refer to each gate in a given time step, and index $i$ to refer to different time steps for the word.

Quantum circuits specify the evolution of the states in discrete time $t \in \mathbb{Z}$. Each time step corresponds to a unitary operation $U_t$, under which the state entering that time step evolves as 
$\mid\psi\rangle_{t}=U_t \mid\psi\rangle_{t-1}$. This time evolution is defined for words in different time steps. 
To separate the words in the quantum circuit, we use the fact that in DisCoCat, different words in a sentence do not have connections other than CNOT gates. The only connection that possibly exists between them is CNOT.

The main steps of the algorithm for processing and encoding the quantum circuit of a sentence are as follows:

\begin{itemize}
    \item \textbf{Step 1: }Partitioning the quantum circuit into words.
\\
    \item \textbf{Step 2: }Scheduling each partition in the quantum circuit into time steps.
\\
    \item \textbf{Step 3: }Arranging gates in different time steps of each partition sequentially from top down.
\\
    \item \textbf{Step 4: } Gate-level encoding of each quantum gate.
\end{itemize}

On the right side of figure \ref{QMT} is a graphical representation of the encoding of a sample quantum circuit for the sentence ``Sara sees Bob'' with five qubits as an input to the LSTM network.

The state $\vert\psi\rangle=\vert{q_0}\rangle\vert{q_1}\rangle\vert{q_2}\rangle\vert{q_3}\rangle\vert{q_4}\rangle$ of the whole sentence is divided into a vector of size equal to the number of words, and so after qubits are partitioned, we have: $P_{qc}=\left[\vert{q_0}\rangle, \vert{q_1}\rangle\vert{q_2}\rangle\vert{q_3}\rangle, \vert{q_4}\rangle\right]$. In addition, after scheduling, each vector is a 7-dimensional array, which is the number of time steps in the quantum circuit in this example.
 
Ignoring the CNOT and without optimising the circuit for the depth, we have seven distinguishable time steps in this example. Since CNOTs connect different words in the quantum circuit, their encoding is not involved in the word encoding. In this quantum circuit, there are three words, and for example, the vector structure of word embedding for the second word is
$E_{w_2} = [t_{w_2}^1, t_{w_2}^2, \cdots, t_{w_2}^7]$.

Then, elements in each 7-dimensional array in the vector have a variable size which depends on the type of the quantum gates in the particular time step. For example, for the third time step of the second word, it is $ t_{w_2}^3 = [[{g_1}^3], [{g_2}^3]] $. 

This encoding makes the decoding part of the network straightforward. LSTM delivers the output in the same format as the encoded quantum circuit.

By processing the input quantum circuit as above, we calculated the encoding of each time evolution for different words in the sentence separately. We applied zero-padding for unitary operations of the words with fewer qubits to have fixed-size inputs to the network.

\subsection{Training classical neural network for language translation via QNLP}

This section describes the classical-quantum encoder-decoder model for the translation task. The encoder reads the sentence to be translated while the decoder translates it into the target language. Our encoding scheme transforms sentences into parameterized quantum circuits (PQCs), with each circuit having two layers. We ensure that synonymous words in different languages have the same angles for their rotation gates, although the circuit structures for different languages may differ. We trained a neural machine translation (NMT) model on the encoded representations of PQCs. The quantum circuit model is at the word level, as is the NMT model. We encode the features of the quantum circuit for each sentence and feed it as an input vector to a conventional Long short-term memory (LSTM) encoder-decoder model. LSTM is a popular choice for $\it{seq2seq}$ translation due to its ability to alleviate vanishing and exploding gradients, making it more effective than the original RNN model for sequence-to-sequence tasks such as machine translation. Finally, the output of the LSTM network is decoded back into the quantum circuit model of the sentence, with the corresponding encoded output of the network.

We utilized the pad$\_$sequences function to ensure that all encoded sequences have the same length. Then, we constructed an encoder-decoder LSTM model for our task.
In this model, each word in the sentence corresponding to the encoded sub-circuit in the quantum circuit is treated as a time step in a sequence.

In the first configuration, both the encoder and decoder consist of LSTM layers with 32 units. To connect these layers, we employed a RepeatVector layer, and finally, we wrapped a dense layer in a time-distributed wrapper. In the second configuration, we used two LSTM layers each with 100 units, and a dense layer with 24 neurons, which corresponds to the dimension of each word in the quantum circuit.

Afterward, we compiled the model using the basic stochastic gradient descent (SGD), RMSprop, and Adam optimizers and sparse categorical cross-entropy loss and trained it on the padded sequences. SGD is an optimization algorithm that aims to minimize a function by employing a fixed learning rate throughout the training process. However, due to its tendency to exhibit high oscillation, increasing the learning rate becomes challenging, resulting in slower convergence. To address this issue, alternative optimizers have been developed. One such optimizer is RMSprop, which tackles the problem by adjusting the learning rate for each parameter individually. It accomplishes this by maintaining an exponentially decaying average of past squared gradients for every parameter. Similarly, the Adam optimizer also adjusts the learning rate for each parameter individually, utilizing the history of gradients. This adaptive approach enables the optimizer to automatically scale the learning rate for each parameter, considering the specific characteristics of the observed gradients during training. In the context of LSTM networks, where distinct parameters may possess varying ranges and properties (e.g., weights and biases of input and forget gates), employing adaptive learning rates can lead to faster convergence and enhanced overall performance.

\begin{figure}[H]
\centering
\includegraphics[scale=0.24]{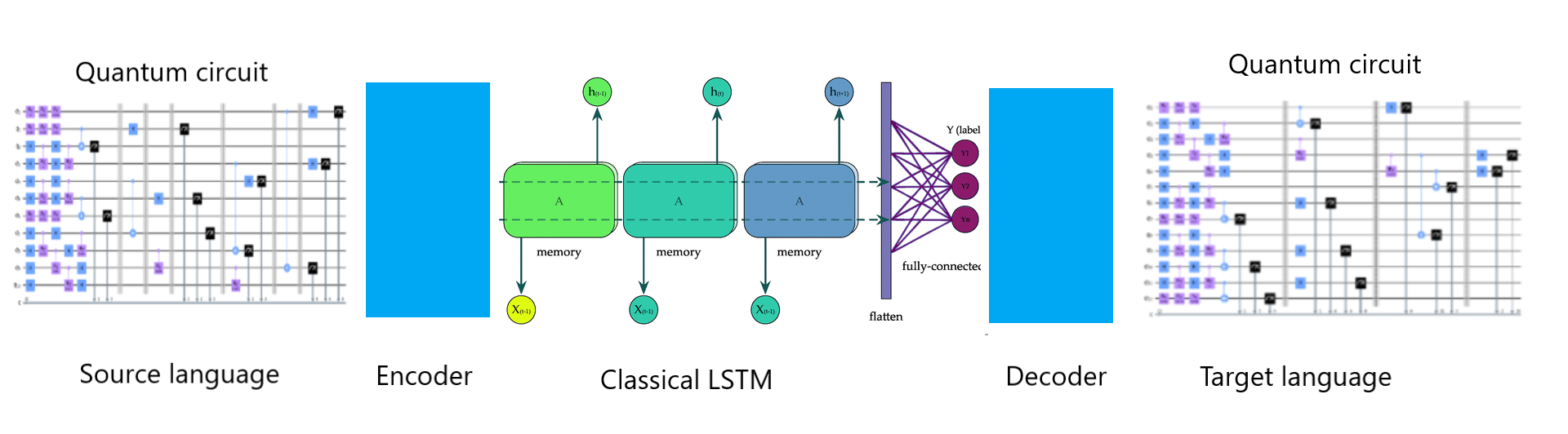}
\captionof{figure}{Architecture of long short-term memory (LSTM) model training.} 
\label{LSTM_Model}
\end{figure}

Our dataset comprised 160 sentences in two languages, English and Persian. The LSTM model for the translation task is illustrated in Figure \ref{LSTM_Model}.

\subsection{Experimental results of implementing the translation task for QNLP}

We utilized the dataset generated in the preceding section consisting of 160 sentences for training our encoder-decoder model. In our implementation, we employed the softmax activation function, a widely utilized choice for the output layer of LSTM-based machine translation models. This activation function serves the purpose of transforming the model's logits or scores into a probability distribution across the target vectors. The output layer of our model employed softmax as the activation function while we used the Adam and RMSprop optimizers and sparse categorical cross-entropy loss during the training process. Our evaluation of the encoded quantum circuit dataset of English-Persian sentences was performed using the trained network.

Table \ref{4} presents the results of various models for quantum machine translation. We trained several LSTM models and report the performance of the top three models. The first model consists of two LSTM layers in the encoder, one LSTM layer in the decoder, and a dense layer with softmax activation. The second model is a basic LSTM model with two LSTM layers and a softmax activation function. The last model combines an LSTM layer with RepeatVector, another LSTM layer, and a dense decoder layer in the output. In this context, the RepeatVector operation serves as a crucial bridge, facilitating the provision of the context vector to the second LSTM layer, which functions as the decoder. Through this mechanism, the RepeatVector operation plays a pivotal role in duplicating and propagating the encoded information from the encoder to the decoder, ensuring its availability for subsequent processing. In our experiment the last model performs the best with the lowest mean absolute and squared errors.

\begin{figure}[H]
\centering
\includegraphics[width=0.8\textwidth]{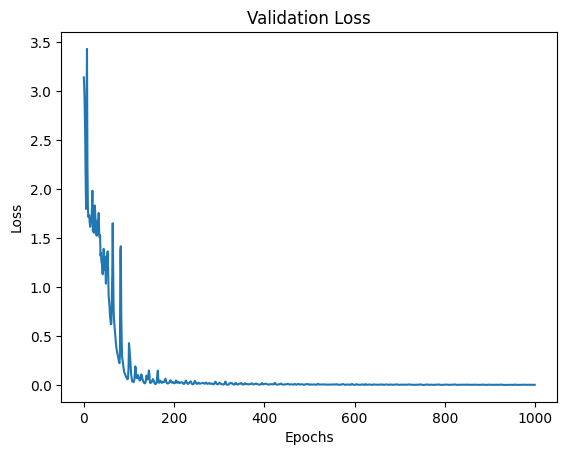}
\captionof{figure}{Validation loss over 1000 epochs for LSTM Model 3 with Adam optimizer.}
\label{LossT3}
\end{figure}

Although our dataset is relatively small for a real-world machine translation task, we aim to showcase the combined architecture of a classical machine translation model with a quantum computer.
Table \ref{5} reports the validation loss for different epoch numbers using SGD, Adam, and RMSprop optimizers for 1000 epochs. 
The convergence of the validation loss over 1000 epochs for the Adam optimizer is shown in Figure \ref{LossT3}. The plot indicates that the loss function converges after approximately 200 epochs.

\begin{table}[h]
\begin{center}
\caption{Experimental results of each LSTM model for the translation task. MAE = mean absolute error; MSE = mean squared error.}\label{4}
\begin{tabular}{llll}

\bfseries{Model} & \bfseries{MAE} & \bfseries{MSE}\\
\hline
Model 1 & 0.8156 & 2.8414 \\

Model 2 & 0.6428 & 0.0021\\

Model 3 & 0.0324 & 0.0019\\

\end{tabular}
\end{center}
\end{table} 

\begin{table}[h]
\begin{center}
\caption{Experimental results of LSTM Model 3---with LSTM layer, RepeatVector, another LSTM layer, and dense decoder layer---for different optimizers.}
\label{5}
\begin{tabular}{ll}

\bfseries{Optimizer} & \bfseries{Loss}\\
\hline
SGD & 0.6423\\

Adam &  0.01574\\

RMSprop & 0.03078 \\

\end{tabular}
\end{center}
\end{table} 

\subsection{A short discussion on complexity}
Quantum natural language processing (QNLP) algorithms have the potential to outperform classical algorithms in certain scenarios, offering significant speedup and improved efficiency for natural language processing (NLP) applications. One advantage of QNLP algorithms is their ability to encode meaning vectors more compactly than classical methods. For instance, eight qubits can store a 256-dimensional classical vector in a quantum system, while 12 qubits can store a 4096-dimensional classical vector. This can lead to faster processing times and lower memory usage, especially when working with large corpora of word embeddings. Moreover, QNLP algorithms have shown quadratic speedup over classical methods for certain tasks. For instance, quantum algorithms like Grover's can perform target pattern-matching searches on unsorted databases of up to $N=2^8=256$ items using only eight qubits in O($\sqrt N$) time, whereas classical algorithms require O($N$) time. This can be advantageous for large datasets and complex language models. Quantum language models (QLMs) are another type of QNLP algorithm that has shown faster convergence and better accuracy than classical models in certain scenarios. For instance, a QLM achieved a speedup of 10 to 50 times over a comparable classical algorithm while maintaining accuracy in one study \cite{?}. Quantum LSTM protocols have also been shown to outperform classical LSTM in certain testing cases, paving the way for QNLP implementation for sequence modeling on NISQ devices. However, much more research and development is needed to fully explore the capabilities of QNLP and its potential impact on the field of NLP. While quantum algorithms for language translation 
are still in their early stages of development, QNLP has the potential to offer significant speedup and improved efficiency over classical algorithms in certain scenarios. 

\section{Conclusion}\label{sec13}
In conclusion, this study aimed to investigate the potential for using quantum natural language processing algorithms on NISQ devices for language translation. By utilizing Shannon entropy, we demonstrated the importance of angles in the performance of parametrized quantum circuits and showed how angles of rotation gates can be used as a means of communication between quantum circuits of different languages. We implemented the translation task using an encoder-decoder model with LSTM and achieved promising results on a small dataset of English and Persian sentences. However, we recognize that more complex machine translation models may be necessary to handle more intricate sentence structures. Our research lays the foundation for future work in quantum neural machine translation, which could offer advantages over classical methods in the future.

\bmhead{Data Availability} \label{Data}
The authors declare that all data supporting the findings of this study are available within the article and its Supplementary Information files. Source data for implementing the translation task and 160 quantum circuits for English and Persian sentences are available at:
\url{https://github.com/Mina-Abbaszade/Data/blob/main/Translate(English).ipynb}
and
\url{https://colab.research.google.com/github/Mina-Abbaszade/Data1/blob/main/Translate(Persian).ipynb}.

The numerical matrix representation of all differences between Shannon entropies of quantum circuits in Table \ref{1} are available at: \url{https://github.com/Mina-Abbaszade/Data/raw/main/Matrix_Representation.ipynb}.


\backmatter

\bmhead{Acknowledgments}
This work was enabled by quantum computing accesses facilitated by the IBM-HBCU Quantum Center at Howard University. We express our appreciation to Dr. Sanjaya Lohani, Manny Gomez, and Dr. Thomas Searles for helpful discussions on running quantum circuits and analyzing the results.




\begin{thebibliography}{9}
	
	\bibitem{Turing} A. Turing, Computing machinery and intelligence, Mind, 236,  433-460, 1950.
	
	\bibitem{Popel} M. Popel, et.al., Transforming machine translation: a deep learning system reaches news translation quality comparable to human professionals,  Nature Communications 11, 4381, 2020.
	
	\bibitem{Anschuetz} E.R. Anschuetz, et. al., Interpretable Quantum Advantage in Neural Sequence Learning, PRX Quantum 4, 020338, 2023.
	
	\bibitem{GPT3} TB. Brown et al., Language models are few-shot learners. arXiv 2005.14165, 2020.
	
	\bibitem{chatGPT} Fuchs, Kevin. Exploring the opportunities and challenges of NLP models in higher education: is Chat GPT a blessing or a curse?. In Frontiers in Education, vol. 8, 1166682. Frontiers, 2023.
	
	\bibitem{Lorenz}
	Lorenz, Robin, et al. QNLP in practice: Running compositional models of meaning on a quantum computer. Journal of Artificial Intelligence Research 76, 1305-1342, 2023.
	
	\bibitem{Shervin}
	Shervin Le Du, Senaida Hernández Santana, Giannicola Scarpa. A gentle introduction to Quantum Natural Language Processing. arXiv:2202.11766, 2022.
	
	\bibitem{Yulin}
	Yulin Wu et al. Strong Quantum Computational Advantage Using a Superconducting Quantum Processor. Physical Review Letters, 127(18):180501, 2021.
	
	\bibitem{NRP2022} Editorial. 40 years of quantum computing. Nature Reviews Physics, 4(1):9-10, 2022.
	
	\bibitem{Supremacy} Arute, Frank, et al. Quantum supremacy using a programmable superconducting processor. Nature 574.7779, 505-510, 2019.
	
	\bibitem{QML} Biamonte, Jacob, et al. Quantum machine learning. Nature 549.7671, 195-202, 2017.
	
	\bibitem{Schuld} M. Schuld, et.al., An introduction to quantum machine learning, Contemporary Physics, 51, 172-185, 2015.
	
	\bibitem{QML2} Schuld, Maria, and Nathan Killoran. Quantum machine learning in feature Hilbert spaces. Physical review letters 122.4, 040504, 2019.
	
	\bibitem{QML3} Schuld, Maria, et al. Circuit-centric quantum classifiers. Physical Review A 101.3, 032308, 2020.
	
	\bibitem{QML4} Preskill, John. Quantum computing in the NISQ era and beyond. Quantum 2, 79, 2018.
	
	\bibitem{QML5} Butler, Keith T., et al. Machine learning for molecular and materials science. Nature 559.7715, 547-555, 2018.
	
	\bibitem{Havlicek} Havlicek, Vojtech, et al. Supervised learning with quantum-enhanced feature spaces. Nature 567.7747, 209-212, 2019.
	
	\bibitem{QML7} Liu, Yunchao, Srinivasan Arunachalam, and Kristan Temme. A rigorous and robust quantum speed-up in supervised machine learning. Nature Physics 17.9, 1013-1017, 2021.
	
	\bibitem{QML8} Huang, Hsin-Yuan, et al. Quantum advantage in learning from experiments. Science 376.6598, 1182-1186, 2022.
	
	\bibitem{QML9} Cong, Iris, Soonwon Choi, and Mikhail D. Lukin. Quantum convolutional neural networks. Nature Physics 15.12, 1273-1278, 2019.
	
	\bibitem{QML10} Daley, Andrew J., et al. Practical quantum advantage in quantum simulation. Nature 607.7920, 667-676, 2022.
	
	\bibitem{QML11} Cerezo, M., et al. Challenges and opportunities in quantum machine learning. Nature Computational Science 2.9, 567-576, 2022.
	
	\bibitem{f} B. Coecke, G. Felice, K. Meichanetzidis, A. Toumi, Foundations for Near-Term Quantum Natural Language Processing, arXiv preprint 2012.03755, 2020.
	
	\bibitem{MLST} L. O'Riordan,  M. Doyle, F. Baruffa and V. Kannan, A hybrid classical-quantum workflow for natural language processing, Mach. Learn. Sci. Technol. 2, 015011, 2020.
	
	\bibitem{picturing} B. Coecke, A. Kissinger, Picturing Quantum Processes, A First Course in Quantum Theory and Diagrammatic Reasoning. Cambridge University Press, Cambridge, 2017.
	
	\bibitem{protocol} S. Abramsky, B. Coecke,
	A Categorical Semantics of Quantum Protocols, in:19th Annual IEEE Symposium on Logic in Computer Science, 415-425, 2004.
	
	\bibitem{math} B. Coecke, M. Sadrzadeh, S. Clark, Mathematical foundations for a Compositional Model of Meaning, Linguistic Analysis, 36, 345-384, 2010.
	
	\bibitem{frob} M. Sadrzadeh, S. Clark, B. Coecke, The Frobenius anatomy of word meaning I: subject and object relative pronouns. Journal of Logic and Computation, 23, 6, 1293-1317, 2013.
	
	\bibitem{inter} B. Coecke, R. Duncan, Interacting quantum observables: categorical algebra and diagrammatics, 13, New J. Phys, 043016, 13, 2011.
	
	\bibitem{Mariam} Salehi, Tahereh, et al. An optimizing method for performance and resource utilization in quantum machine learning circuits. Scientific Reports, 12.1, 16949, 2022.
	
	\bibitem{circuit} K. Meichanetzidis, S. Gogioso, G. D. Felice, N. Chiappori, A. Toumi and B. Coecke, Quantum natural language processing on near-term quantum computers,arXiv:2005.04147, 2020.
	
	\bibitem{lambeq} D. Kartsaklis, et. al., lambeq: An Efficient High-Level Python Library for Quantum NLP, arXiv 2110.04236, 2021.
	
	\bibitem{Mina} M. Abbaszade, et. al., Application of quantum natural language processing for language translation, IEEE Access, 9, 130434-130448, 2021.
	
	\bibitem{thesis} Johanna Monti. Multi-Word Unit Processing in
	Machine Translation. PhD thesis, Universita Degli Studi Di Salerno, 2013.  
	
	\bibitem{Towards Machine translation} I. Vicente Nieto, Towards Machine translation with Quantum Computers. Master’s Degree Project, Physics. Stockholm, Sweden, 2021. 
	
	\bibitem{R} R. Narayan, et.al., Quantum neural network based machine translator for English to Hindi, Applied Soft Computing, 38, 1060-1075, 2016.
	
	\bibitem{M.Schuld} M. Schuld, et.al.,
	The quest for a quantum neural network,
	Quantum Information Processing 13, 2567-2586, 2014.
	
	\bibitem{Peral} D. Peral García, et.al., Systematic Literature Review: Quantum Machine Learning and its applications, arXiv:2201.04093, 2022.
	
	\bibitem{Caro} M.C. Caro, et.al.,
	Generalization in quantum machine learning from few training data, Nature Communications 13, 4919, 2022.
	
	\bibitem{match} Jung, Heeseung, et al. Impact of Sentence Representation Matching in Neural Machine Translation. Applied Sciences 12.3, 1313, 2022.
	
	\bibitem{p} B. Coecke, Quantum Picturalism, Contemporary Physics, 51, 59-83, 2010.
	
	\bibitem{Grefenstette} E. Grefenstette, et. al., Concrete Sentence Spaces for Compositional Distributional Models of Meaning, Proceedings of the 9th International Conference on Computational Semantics (IWCS11), 2011.
	
	\bibitem{m} M. Sadrzadeh, Pregroup Analysis of Persian Sentences, In C. Casadio and J. Lambek, editors, Computational algebraic approaches to natural language, Polimetrica, 2006.
	
	\bibitem{algebraicapproach} J. Lambek and C.Casadio, editors, Computational algebraic approaches to natural language,Polimetrica, (2006).
	
\end{thebibliography}
\end{document}